\documentclass{article}

\usepackage[final, nonatbib]{neurips_2020}

\usepackage[utf8]{inputenc} % allow utf-8 input
\usepackage[T1]{fontenc}    % use 8-bit T1 fonts
\usepackage{hyperref}       % hyperlinks
\usepackage{url}            % simple URL typesetting
\usepackage{booktabs}       % professional-quality tables
\usepackage{amsfonts}       % blackboard math symbols
\usepackage{nicefrac}       % compact symbols for 1/2, etc.
\usepackage{microtype}      % microtypography

\usepackage{graphicx}
\usepackage{subfig}
\usepackage[dvipsnames]{xcolor}

\title{Neural language models for text classification in evidence-based medicine}

\author{%
  Andrés Carvallo,
  Denis Parra\\
  Department of Computer Science, Pontificia Universidad Católica de Chile, Santiago, Chile\\
  \texttt{afcarvallo@uc.cl}, \texttt{dparra@ing.puc.cl} \\
  \And
  Gabriel Rada,
  Daniel Pérez,
  Juan Ignacio Vásquez,
  Camilo Vergara \\
  Epistemonikos Foundation, Santiago, Chile \\
  \texttt{\{radagabriel,dperezrada,juan,camilo\}@epistemonikos.org}
}

\begin{document}

\maketitle

\begin{abstract}

The COVID-19 has brought about a significant challenge to the whole of humanity, but with a special burden upon the medical community. Clinicians must keep updated continuously about symptoms, diagnoses, and effectiveness of emergent treatments under a never-ending flood of scientific literature. In this context, the role of evidence-based medicine (EBM) for curating the most substantial evidence to support public health and clinical practice turns essential but is being challenged as never before due to the high volume of research articles published and pre-prints posted daily. Artificial Intelligence can have a crucial role in this situation. In this article, we report the results of an applied research project to classify scientific articles to support Epistemonikos, one of the most active foundations worldwide conducting EBM. We test several methods, and the best one, based on the XLNet neural language model, improves the current approach by 93\% on average F1-score, saving valuable time from physicians who volunteer to curate COVID-19 research articles manually.

\end{abstract}

\section{Introduction}
Evidence-based medicine (EBM) is a medical practice that aims to find all the evidence to support medical decisions. This evidence nowadays is obtained from biomedical journals, usually accessible through online databases like PubMed\cite{lindsey2013pubmed} and EMBASE\cite{lefebvre2008enhancing}, which provide free access to articles' abstracts and in some cases, to full articles. In the context of the COVID-19 pandemic, EBM is critical to making decisions at the individual level and public health since research articles address topics like treatments, adverse cases, and effects of public policies in medicine. The EBM foundation Epistemonikos has made essential contributions by curating and publishing updated guides of what treatments are working and not against COVID-19~\footnote{http://epistemonikos.org/}. Epistemonikos addresses EBM by a combination of software tools for data collection, storage, filtering \cite{donoso2018interactive,carvallo2020automatic}, and retrieval, as well as by the vital labor of volunteer physicians who curate and label research articles based on quality (to include in the database), type (systematic review, randomized trial, among others) and PICO labels (patient, intervention, comparison, outcome). However, this workflow has been challenged during 2020 by increasing growth and rapidly evolving evidence of COVID-19 articles published in the latest months. Moreover, to ensure the rapid collection of the latest evidence published, pre-print repositories such as medRXiv and bioRXiv have been added to the traditional online databases.
In order to support Epistemonikos' effort to filter and curate the flood of articles related to COVID-19, we present the results of an applied AI project where we implement and evaluate a text classification system to filter and categorize research articles related to COVID-19. The current model, based on Random Forests, has an acceptable performance classifying systematic reviews (SR) but fails on classifying other document categories. In this article, we show how using BioBERT yields marginal improvements, while XLNET results in significant progress with the best performance. These results save a considerable amount of time from volunteer physicians by pre-filtering the articles worth of manual curation and labeling for EBM. In average, a physician takes two minutes in reviewing one article, while the system we present in this article can review up to $32,000$ within one hour.

%With the help of volunteer physicians, they classify emergent literature for the COVID-19 virus in systematic reviews, broad syntheses, or primary studies, which is the first step for finding relevant clinical evidence. Until now, they produced a Random Forest model for classifying documents into different categories. However, in this paper, we show how the use of Transformers-based Language Models (XLNET) helped this foundation save significant effort to their collaborators.

%\section{Related Work}

\section{Methods and results}

%\subsection{Methods and data}
\textbf{Methods and data}. We compare document classification results among a (i) random forest with a customized tokenizer made by Epistemonikos, (ii) an XLNET \cite{yang2019xlnet} language model representing documents using a linear layer as a classifier, and (iii) the same setting with a BioBERT \cite{lee2020biobert}language model. 
The documents' classification can be a systematic review, a primary study using a randomized controlled trial, non-randomized primary study, broad synthesis, and excluded document.  The distribution of documents can be observed in the second column of Table \ref{metrics_results}.  Notice that the type of document partially explains the classification models' mistakes: broad synthesis and systematic review are both kinds of surveys, while primary studies (rct and non-rct) deal with specific treatments and populations. Excluded can be of any of the other four classes, but they are not included in the official Epistemonikos dataset due to their low quality.

% \begin{table}[htb]
%     \caption{Distribution of documents per class}
%     \centering
%     \begin{tabular}{c|c}
%     \toprule
%     Document class & \# documents \\
%     \hline
%     Systematic review  & 286,050  \\
%     Primary rct  & 56,623  \\
%     Primary not-rct  & 35,644  \\
%     Broad synthesis  & 17,324 \\
%     Excluded &  6,096\\
%     \bottomrule
%     \end{tabular}
%     \label{dataset}
% \end{table}

%\subsection{Results}
\textbf{Results}. Table \ref{metrics_results} shows the performance of each model in terms of precision (Prec.), recall (Rec.), and F1-score (F-1) for every type of document.
% In general
In general terms, we observe that XLNet obtains the top F-1 score for any document category, in some cases by a small margin, such as under systematic review (F-1=$.97$), and in other cases by a large margin, as in the classes Broad synthesis (F-1=$.61$), and Excluded (F-1=$.78$). The results indicate that the random forest and BioBERT with a linear layer have a bias towards the most dominant class, \textit{Systematic review}, reporting slightly better recall (Rec.=$.99$ and Rec.=$1.0$) than XLNet (Rec.=$.98$) in this particular type of document. However, XLNet is better than the other two models in terms of Precision upon all classes, with the only exception of \textit{Broad synthesis}, where random forest (Prec.=$.75$) performs better than XLNet (Prec.=$.67$). However, XLNet recall outperforms (Rec.=$.56$) random forest (Rec.=$.15$).
%The results show an improvement of XLNET for each class's other models in terms of recall, except perfect recall for SR in BioBERT. Moreover, for precision, we observe a similar behavior, but XLNET surpasses all the different methods for each type of document. Finally for f1-score, which measures recall and precision we observe a considerable improvement over the other methods, specially for the minority classes (BS and exc). 
It is important to note that when using the random forest implemented for Epistemonikos, a new tokenizer has to be made depending on the document categories. In the case of XLNET, it is more versatile because it is enough to train embeddings and classify them regardless of the document category. In the case of BioBERT, which has a similar operation, it does not yield consistent performance for the minority classes \textit{Broad synthesis} and \textit{Excluded}.

% Please add the following required packages to your document preamble:
% \usepackage{booktabs}
\begin{table}[htb]
\caption{Distribution of document and results obtained for document classification of Broad Synthesis, Systematic Review, Primary Study randomized controlled trial (Primary rct), Primary Study non-randomized controlled trial (Primary non-rct), and Excluded.}

\centering
\small
\begin{tabular}{@{}lc|ccc|ccc|ccc@{}}
\toprule
               &   & \multicolumn{3}{c|}{Random Forest} & \multicolumn{3}{c|}{XLNet}   & \multicolumn{3}{c}{BioBERT} \\ \midrule
                & \# docs.  & Prec.      & Rec.      & F-1      & Prec. & Rec. & F-1          & Prec.    & Rec.    & F-1    \\
                  \midrule
Broad synthesis  & 17,324 & .75        & .15       & .26      & .67   & .56  & \textbf{.61} & 0        & 0       & 0      \\
Systematic review & 286,050 & .93        & .99       & .96      & .96   & .98  & \textbf{.97} & .85      & 1.0     & .92    \\
Primary rct      & 56,623 & .25        & .79       & .38      & .94   & .85  & \textbf{.89} & .71      & .71     & .71    \\
Primary non-rct   & 35,644 & .63        & .40       & .49      & .64   & .91  & \textbf{.75} & .61      & .90     & .72    \\
Excluded          & 6,096 & .70        & .21       & .32      & .82   & .74  & \textbf{.78} & 0        & 0       & 0      \\ \bottomrule
\end{tabular}
\label{metrics_results}
\end{table}

% \begin{table}[htb]
% \centering
% \small
% \begin{tabular}{@{}c|c|ccccc}
% \textbf{Model} & \textbf{Metric} & {\textbf{Precision}} & {\textbf{Recall}} & {\textbf{F1-Score}} 
% \\ \hline
%  & BS & .75 & .15 & .26     \\
%  & Exc & .70 & .21 & .32\\ 
%  Random Forest & PS-not-RCT & .63  & .40  & .49   \\
%  & PS-RCT & .25 & .79 & .38  \\
%  & SR & .93 & .99 & .96  \\
%  \hline
 
%  & BS & .67 & .56 & .61    \\
%  & Exc & .82 & .74  & .78    \\ 
%  XLNET & PS-not-RCT & .64 & .91 & .75  \\
%  & PS-RCT & .94 & .85 & .89  \\
%  & SR & .96 & .98 & .97  \\
%  \hline
 
%  & BS & 0 & 0 & 0    \\
%  & Exc & 0 & 0 & 0\\ 
%  BioBERT & PS-not-RCT & .61 & .90 & .72  \\
%  & PS-RCT & .71 & .71 & .71  \\
%  & SR & .85 & 1.0 & .92  \\
%  \hline
 
% \end{tabular}

% \caption{Results obtained for document classification in Broad Synthesis (BS) , Excluded (Exc) , Primary Study non randomized controlled trial (PS-non-rct) , Primary Study randomized controlled trial (PS- rct) and Systematic Review (SR). Using precision, recall and f1-score for each of the compared models, Random Forest, XLNET and BioBERT.}
% \label{metrics_results}
% \end{table}

\section{Conclusion}
In this study, we have compared three methods, one of which is currently in production at the Epistemonikos foundation, the random forest. The others are BioBERT, which, although it is based on the transformer architecture, does not achieve the results shown by XLNET. Having such reliable results means a big impact in times of the COVID-19 pandemic where there is an exponential growth of available literature. In future work we will incorporate explanations obtained from transformer attention mechanisms, compare them against other explanation methods like LIME\cite{ribeiro2016should} or SHAP\cite{lundberg2017unified}, and conduct a user study to assess whether physicians' work is facilitated by this feature.

\section*{Broader Impact}
This work seeks to decrease manual effort in the practice of evidence-based medicine, allowing physicians to distinguish relevant documents for clinical questions. Implementing the method with the largest performance in our offline evaluation (XLNet) in production might imply an increased cost in terms of GPU needs for Epistemonikos, which is not under their current infrastructure. Adding more documents might also imply additional fine-tuning of the model, incurring in larger costs. Another aspect not addressed in this research is that of fairness: is the current model performing better to classify certain populations being treated (e.g. white males)  compared to black females? we should address this aspect actively to prevent our model from learning undesired biases already seen in several applications. %However, improvement is seen using the XLNET model, which is currently in production by Epistemonikos. A potential discussion is to incorporate explanations obtained from transformer attention mechanisms and see if physicians consider this.

\bibliographystyle{plain}
{\small
\bibliography{references.bib}

\begin{thebibliography}{1}

\bibitem{carvallo2020automatic}
Andres Carvallo, Denis Parra, Hans Lobel, and Alvaro Soto.
\newblock Automatic document screening of medical literature using word and
  text embeddings in an active learning setting.
\newblock {\em Scientometrics}, pages 1--38, 2020.

\bibitem{donoso2018interactive}
Ivania Donoso-Guzm{\'a}n and Denis Parra.
\newblock An interactive relevance feedback interface for evidence-based health
  care.
\newblock In {\em 23rd International Conference on Intelligent User
  Interfaces}, pages 103--114, 2018.

\bibitem{lee2020biobert}
Jinhyuk Lee, Wonjin Yoon, Sungdong Kim, Donghyeon Kim, Sunkyu Kim, Chan~Ho So,
  and Jaewoo Kang.
\newblock Biobert: a pre-trained biomedical language representation model for
  biomedical text mining.
\newblock {\em Bioinformatics}, 36(4):1234--1240, 2020.

\bibitem{lefebvre2008enhancing}
Carol Lefebvre, Anne Eisinga, Steve McDonald, and Nina Paul.
\newblock Enhancing access to reports of randomized trials published
  world-wide--the contribution of embase records to the cochrane central
  register of controlled trials (central) in the cochrane library.
\newblock {\em Emerging Themes in Epidemiology}, 5(1):13, 2008.

\bibitem{lindsey2013pubmed}
Wesley~T Lindsey and Bernie~R Olin.
\newblock Pubmed searches: Overview and strategies for clinicians.
\newblock {\em Nutrition in Clinical Practice}, 28(2):165--176, 2013.

\bibitem{lundberg2017unified}
Scott~M Lundberg and Su-In Lee.
\newblock A unified approach to interpreting model predictions.
\newblock In {\em Advances in neural information processing systems}, pages
  4765--4774, 2017.

\bibitem{ribeiro2016should}
Marco~Tulio Ribeiro, Sameer Singh, and Carlos Guestrin.
\newblock " why should i trust you?" explaining the predictions of any
  classifier.
\newblock In {\em Proceedings of the 22nd ACM SIGKDD international conference
  on knowledge discovery and data mining}, pages 1135--1144, 2016.

\bibitem{yang2019xlnet}
Zhilin Yang, Zihang Dai, Yiming Yang, Jaime Carbonell, Russ~R Salakhutdinov,
  and Quoc~V Le.
\newblock Xlnet: Generalized autoregressive pretraining for language
  understanding.
\newblock In {\em Advances in neural information processing systems}, pages
  5753--5763, 2019.

\end{thebibliography}
}

\end{document}